# Collaborative Evolution of 3D Models

**Juan C Quiroz**
Independent contractor, Reno, NV, USA

**Amit Banerjee**
Pennsylvania State University, Harrisburg, PA, USA

**Sushil J. Louis and Sergiu M. Dascalu**
University of Nevada, Reno, NV, USA

We present a computational model of creative design based on collaborative interactive genetic algorithms. In our model, designers individually guide interactive genetic algorithms (IGAs) to generate and explore potential design solutions quickly. Collaboration is supported by allowing designers to share solutions amongst each other while using IGAs, with the sharing of solutions adding variables to the search space. We present experiments on 3D modeling as a case study, with designers creating model transformations individually and collaboratively. The transformations were evaluated by participants in surveys and results show that individual and collaborative models were considered equally creative. However, the use of our collaborative IGAs model materially changes resulting designs compared to individual IGAs.

## Introduction

Design is a goal-oriented, constrained decision-making activity, involving learning about emerging features [1]. It is usually characterized by four phases – conceptual design, detailed design, evaluation, and iterative redesign [2]. Within this context, creativity has the potential to occur when a designer purposely shifts the focus of the search space [3]. Specifically, the ability to perform goal-oriented shifts while brainstorming and exploring potential solutions is crucial to creativity in the design process. This is accomplished implicitly by the designer's understanding of the problem changing over time, or explicitly by considering additional traits which may yield interesting solutions [3].

We present a computational model of creative design based on collaborative interactive genetic algorithms (IGAs) which supports goal-oriented shifts by adding variables to the search space. In our model, designers individually guide IGAs to generate and explore design solutions quickly [2]. The model allows designers to share solutions amongst each other by presenting designers with a sample of solutions generated by their peers. When a designer selects a solution from a peer, the solution is injected into his/her IGA population, with this injection adding new variables to the search space.

In our previous work [4], [5], we introduced our model of creative design and presented a pretest of the model for user guided design of floorplans, but without expanding the search space with additional variables. The pretest results showed that floorplans created collaboratively were considered to be more original than floorplans created individually. Following the pretest, we conducted a user study where we addressed the question of whether collaboration alone–without expanding the search space—introduced the potential to generate creative solutions [5]. Results showed that floorplans created collaboratively were considered to be more revolutionary and original than floorplans created individually.

In this paper, we present experiments where the search space is expanded by adding variables during the evolutionary search. We use 3D modeling as the case study for the experiments. However, rather than creating 3D models from scratch, we explore transformations of 3D models with vertex programs.

The rest of the paper is organized as follows. We start by presenting a discussion of the computational model and the 3D modeling case study. Next, the experimental setup is presented in detail, followed by user study results and discussions.

## Computational Model of Creative Design

Models of creative design presented by the research design community manipulate the search space through the use of techniques, including combination, analogies, transformation, emergence, and first principles [6]. Genetic algorithms provide a means for exploring a search space consisting of potential solutions that meet a given set of requirements [7]. However, there are times when it is difficult, if not impossible to define a suitable fitness function, especially when dealing with problems require subjective evaluation. An interactive genetic algorithm empowers the user to drive evolution by replacing the fitness evaluation [8], and enables users to guide evolution based on their sense of aesthetics, intuition, and domain expertise.

Our computational model of creative design leverages the exploration power of the GA, the visualization and subjective feedback integration of the IGA, and collaboration in order to allow designers to shift the focus of the search space during an evolutionary run. Our model is unique in (1) using IGAs to guide the subjective exploration of changing search spaces, and (2) using collaboration to change the search space by adding variables. By using IGAs in our model, users can incorporate their personal preference, sense of aesthetics, intuition, and expertise into the search process. In addition, each user decides when to take solutions from peers, meaning that the user always remains in control of his/her own IGA.

Collaboration allows designers to share expertise, to be exposed to traits they may not have considered, and to complement each other in the task of exploring solutions which meet a given set of requirements. In our model, designers start with different variable sets. The designers are exposed to solutions being explored by their peers during collaboration and consequently to the different effects resulting from different variable sets. Taking solutions from peers allows designers to expand their search space by automatically incorporating the variables being explored by their peers.

Figure 1 illustrates our computational model of creative design. The figure illustrates three users collaborating with each other. Each user interacts with a GA by acting as the subjective evaluation. Evaluation may consist of subjective evaluation only, or a combination of subjective and objective evaluations. The arrows between the IGAs represent the communication that takes place between the peers. If a user likes a design solution from one of his/her peers, then the user has the option to inject that solution into his/her population, thus introducing a search bias. For implementation details of the collaborative IGA model, see [4] and [5].

Our collaborative IGA computational model is a special case of a case injected genetic algorithm (CIGAR) [9], where (1) each user serves as a case base to peers, and (2) each user determines when and how many individuals to inject into his/her population, instead of injection being done in an algorithmic fashion. When a user chooses to inject a solution from one of his/her peers, the introduced bias will not only become apparent in the user's own population, but it will also be visible to his/her peers, since users can always see a subset of each others' solutions. For example, as "user A" interacts with the IGA, the changes in the population of user A will be reflected on the screens of the peers of user A. Thus, each user participating in collaboration serves as a dynamic case base to his/her peers.

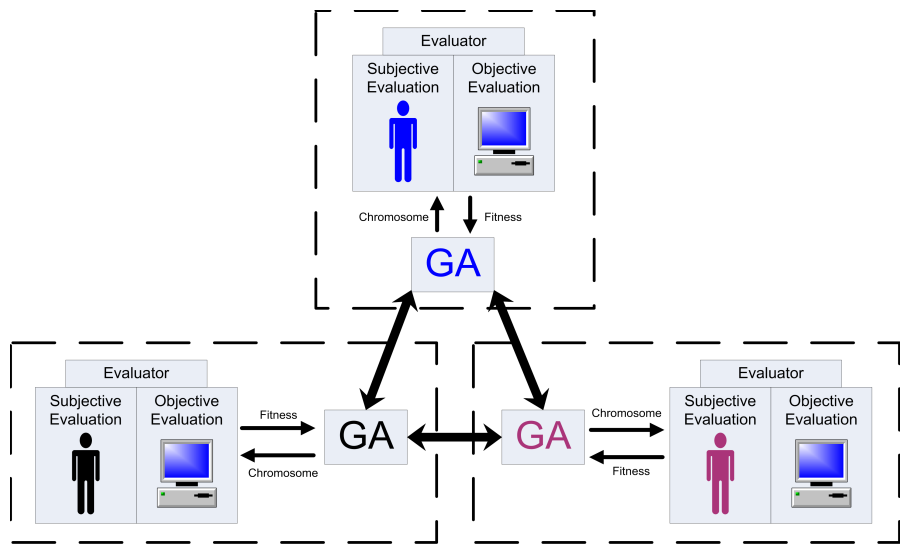

**Fig. 1** Computational model of creative design

Typically in CIGARs a case base of solutions to previously solved problems is maintained. Based on problem similarity, individuals similar to the best individuals in the current population are periodically injected from the case base, replacing the worst individuals in the population [9]. In our computational model, the designer plays the role of determining how many, when, and which individuals to inject at any step during the collaborative evolutionary process. If the injected individuals make a positive contribution to the overall population, then they will continue to reproduce and live on, while injected individuals that do not improve the population performance will eventually die off. Hence, the user is not penalized for injecting subpar individuals. We use fitness biasing (linear scaling) to ensure that injected individuals survive long enough to leave a mark on the host population.

**Creative Potential of Model**

Boden describes two types of creativity in design, P-creativity and H-creativity [10]. P-creativity (personal or psychological creativity) occurs when the design is creative to the designer. H-creativity occurs when the design is creative when compared to all that has been created and produced historically by all of humanity. S-creativity (situated creativity), a third type which has also been presented, occurs when the resulting design is novel to that particular situation, but not necessarily be creative to the individual or creative historically [11]. In our model, at the individual level designers guide P-creative processes while interacting with the IGA. During collaboration, the sharing of design solutions allows the designers as a group to guide the S-creative process. Specifically, users can begin exploration of distinct search spaces (defined by different variable sets), and through collaboration, explore search spaces defined by combinations of their variable sets.

**3D Modeling Representation**

We use 3D modeling as the case study for our experiments. Rather than creating 3D models from scratch, we perform modifications to existing and well-formed 3D models by evolving vertex programs. The vertex programs allow for an operation to be applied on a per vertex basis for every vertex on a 3D model. For the experiments in this paper we used the OGRE 3D rendering engine and Cg as the GPU programming language for the vertex programs.

We evolve vertex programs with genetic programming (GP) [12]. GP is an evolutionary computation technique where each individual in the population is a computer program. The computer program is represented using a tree structure (GP tree) and the operations of the GP tree are typically mathematical operations.

Figure 2 illustrates our representation of the vertex programs in the IGA as a bit encoded binary tree. A binary tree is a tree data structure in which each node has at most two child nodes. For example, a node with index $i$ would have its children at indices $2i + 1$ and $2i + 2$. In our representation, all leaves are at the same depth and every parent node has two child nodes. From the perspective of GA encoding, storing a binary tree in an array has the advantage of being readily mapped to a bit string.

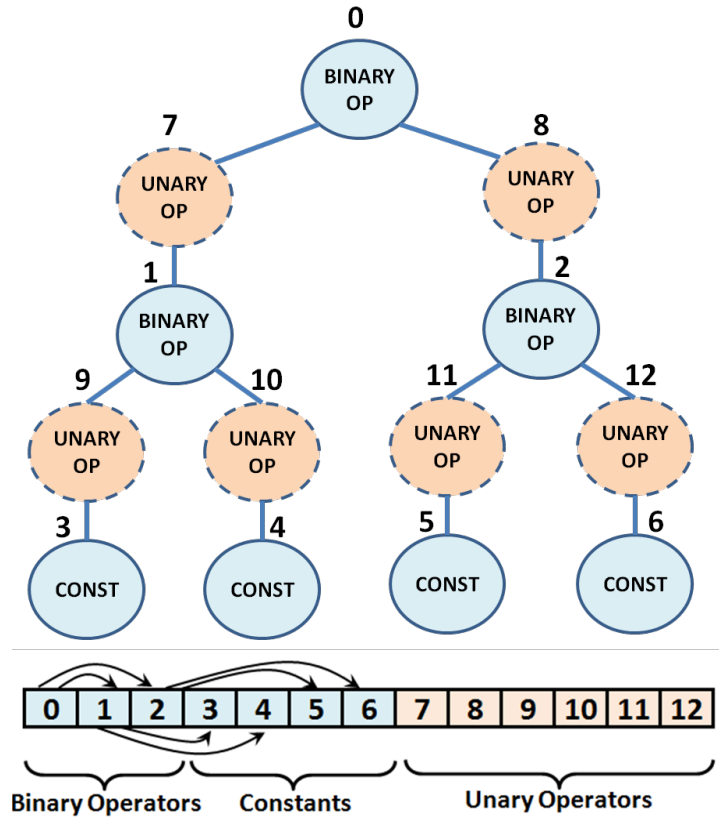

**Fig. 2** Binary string representation of vertex programs.

In our implementation, parent nodes consist of binary operations on the children nodes, with these operations including addition, subtraction, multiplication, and division. It was seen in preliminary experiments that unary operators, such as the trigonometric functions, also created a wide range of interesting transformations of 3D models, and therefore a second array is used to store unary operators applied to each child node in the tree, as shown in Figure 2. We use the trigonometric functions of sine, cosine, and tangent. The leaf nodes in the tree consist of variables including the coordinates of the current vertex (x, y, or z), the current simulation time looping from 0 to $2\pi$, and random numbers between -10 and 10. The binary operations, the set of constants, and the unary operators are combined into a single bit string array manipulated by the IGA as illustrated in Figure 2.

The decoded equation of each individual in the population is written to a Cg file. The decoded equation modifies all of the x, y, and z coordinates with the same equation, or only one of the x, y, or z coordinates. For example, equation (1) modifies the x, y, and z coordinates of each vertex with the same equation:

$$p.xyz = p.xyz + (2.2 - (p.x/11)) + (7 * \cos(p.y)); \quad (1)$$

where p.xyz represents the x, y, and z coordinates of the current vertex, p.x is the x coordinate of the current vertex, and p.y is the y coordinate of the current vertex. The value of the equation on the right is added to each of the x, y, and z coordinates of the current vertex. Thus, equation (1) is equivalent to the following:

$$p.x = p.x + (2.2 - (p.x/11)) + (7 * \cos(p.y));$$
$$p.y = p.y + (2.2 - (p.x/11)) + (7 * \cos(p.y));$$
$$p.z = p.z + (2.2 - (p.x/11)) + (7 * \cos(p.y));$$

The vertex program representation enables us to have a first set of users evolve programs that modify the x coordinate, and a second set of users evolve programs that modify the y coordinate, as illustrated in Figure 3. Through collaboration, the first set of users can inject solutions from the second set of users, resulting in their respective search spaces expanding from exploring equations that only modify the x and y coordinate, or only the y and z coordinate, to equations that modify all the coordinates of the 3D model.

**Experimental Setup**

The experiments presented in this paper were conducted in an environment built with the OGRE rendering engine. The goal of the experiments is to show that the use of our collaborative computational model results in solutions that are more creative compared to solutions generated using a simple IGA. To this end, the experiment consisted of three phases: (1) creation of designs, (2) first evaluation of the designs, and (3) online evaluation of the designs. These are described in detail in the next subsections.

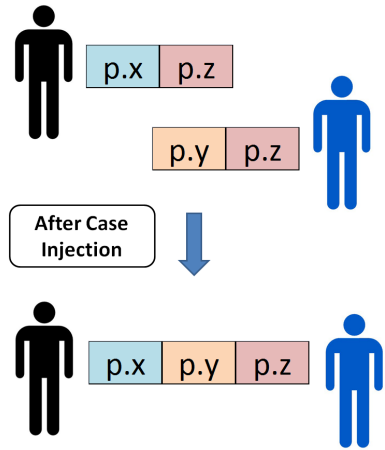

**Fig. 3** Expanding design variable space through collaboration.

### Design Creation

A group of 20 students from the Computer Science Department at the University of Nevada, Reno, ("the design participants") used simple IGAs and our collaborative model to create transformations of the 3D models individually and collaboratively. During the creation of the designs, participants were split into pairs. When using our collaborative model, participants were only allowed to share solutions within each pair. During collaboration between users, the search space was expanded as shown in Figure 3.

We used an ABA experimental design to test the hypothesis that our collaborative model results in more creative solutions when compared to a simple IGA. In the experimental design, the baseline condition (A) was participants creating solutions individually with a simple IGA, and the experimental condition (B) was participants creating solutions collaboratively with our computational model of creative design. The experiment consisted of each design participant conducting an ABA session (individual session, collaborative session, individual session) followed by a BAB session (collaborative session, individual session, collaborative session). The goal of this design is to show that the use of our computational model, rather than time, is the controlling variable if there is a change in behavior between baseline and experimental conditions [13]. Our hypothesis is that the resulting scores (from a 7-point Likert scale) would resemble a zig-zag pattern as illustrated in Figure 4, with the average scores of models gener-ated during individual trials being close to 7, and the average scores of models generated during collaborative trials being close to 1.

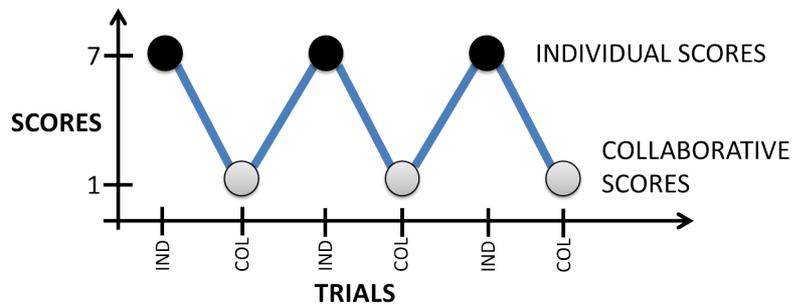

**Fig. 4** Hypothesized comparison of scores between individual and collaboratively created models.

Figure 5 shows the three models used in the user study: (1) a futuristic female model in a blue suit, (2) a green ninja, and (3) a white robot. Half of the design participants were assigned to a first design group, while the other half where assigned to a second design group. The first design group created transformations for the models in this order: (1) **individual** – female, (2) **collaborative** – ninja, (3) **individual** – robot, (ABA) and (4) **collaborative** – robot, (5) **individual** – ninja, and (6) **collaborative** – female (BAB). The second design group created transformations similarly to the first design group, except that in trials 4-6 the second design group begins with the female model instead of the robot model: (4) **collaborative – female**, (5) individual – ninja, and (6) **collaborative robot** (BAB).

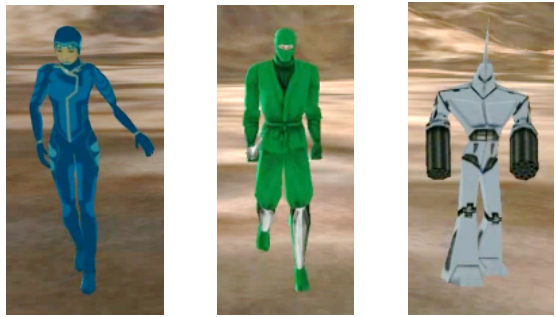

**Fig. 5** Original 3D models used during experiments

Each design participant created at least two transformations for each of the models during the collaborative and the individual sessions. The design participants were then asked to pick the solutions they considered the most creative from each set of models, for a total of six final solutions from each design participant: (1) female - individual, (2) female - collaborative, (3) ninja - individual, (4) ninja - collaborative, (5) robot - individual, and (6) robot - collaborative. This set of final solutions was evaluated as described in the next subsection.

**Evaluation**

Our evaluation is based on the work of Thang et al. [14], which consists of criteria derived from the Creative Product Semantic Scale [14], [15] and [16]. The participants were asked to rate model transformations on a 7-point Likert scale on the following criteria: creativity, novelty, surprising, workability, relevance, and thoroughness. However, instead of simply using the terms from this rating scale, we presented the users with five statements, and participants specified the degree to which they agreed or disagreed with the statements. The evaluation participants were only informed that a group of students had created a set of transformations of models, representing special effects for a video game.

The first evaluation statement, related to creativity, was: "The transformation is creative." Many definitions of creativity exist, and the definition largely depends on the context and problem domain. Therefore, we provided the evaluation participants with the following definition to evaluate the created designs: "A creative transformation is a transformation that is new, unexpected, and valuable." The statement could be answered by selecting between "Extremely Creative" (coded as 1) versus "Not Creative At All" (coded as 7), or with a number in between 1 and 7.

The rest of the evaluation statements did not use the terms workability, relevance, and thoroughness, to avoid ambiguity regarding the meaning of these terms. Instead, the evaluation asked whether the transformation with or without tweaks could be used in a video game, which addressed workability, relevance, and thoroughness. The other four statements in the evaluation were: (1) The transformation can be used in a video game; (2) The transformation with minor tweaks can be used in a video game; (3) The transformation is novel; and (4) The transformation is surprising. Users could answer these questions on a 7-point Likert scale as "Very True" (coded as 1) versus "Not True At All" (coded as 7).

The first group of evaluation participants evaluated the final solution sets created and selected by the design participants. Each evaluation partic-ipant evaluated two final solution sets; hence, each evaluation participant evaluated a total of 12 models.

After the first evaluation, we conducted a second online evaluation. A total of 16 adult volunteers completed the online evaluation. We selected the best six individually created models and the best six collaboratively created models for online evaluation. We made a video of these 12 models, posted the video online, and collected data via an online survey. The online survey used the same evaluation criteria as the first evaluation phase, except that we removed the statement asking whether the transformation with minor tweaks could be used in a video game to make the survey shorter.

## Results and Discussion

Below we present the results of the first evaluation and the online evaluation of the designs created by the design participants. In addition, we provide examples of models created individually and collaboratively, and a discussion of how our collaborative model changes design.

### First Evaluation

Figure 6 illustrates the evaluation scores received from the first evaluation participants for the statement "The transformation is creative." The average scores and boxplots are for the models created by the first design group and the second design group during individual and collaborative trials. The difference in these design groups is the order in which the models were evolved, as shown in the top axis of the plot. The boxplots compare the distributions of scores between individual and collaborative trials. The average scores between individually and collaborative trials were compared using a Student's t-test to verify statistical significance. We did not account for sample size in the statistical analysis.

For the first design group, trials 1-3 exhibit a zig-zag pattern that is the opposite of our hypothesis. We expected the individually created models to receive scores closer to 7 and the collaboratively created models closer to 1, yet the average scores and boxplots show the opposite for trials 1-3. Thus, when collaboration was introduced, the resulting models were considered less creative. For trials 4-6 of the first design group, we see scores supporting our hypothesis. That is, when users created models collaboratively, the resulting models were more creative. For the second design group, only trials 4-6 seem to support our hypothesis that models created collaboratively are more creative.

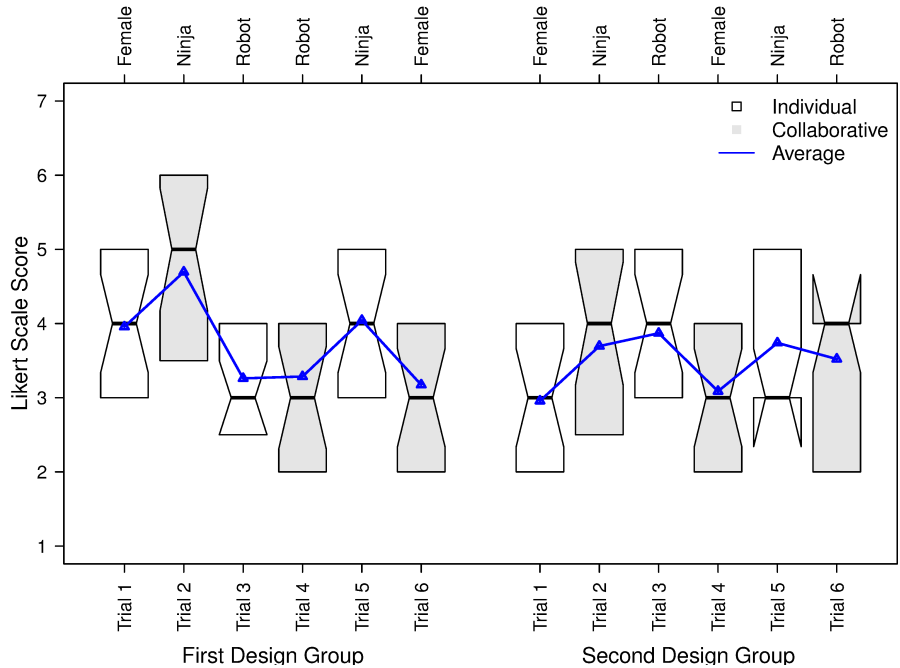

**Fig. 6** Scores for "The transformation is creative" statement (1 – Extremely creative, 7 – Not creative at all).

Using the Student's t-test, we compared the average scores between successive trials to determine whether the changes between the average scores of collaborative and individually created models were statistically significant. For the first design group, the change from trial 2 (collaborative) to trial 3 (individual) was statistically significant ($p < .05$). This was unexpected because it shows that when users created models individually, after having created models collaboratively, the resulting models were scored as more creative. This is also clear from the lack of overlap between the confidence intervals of the medians (denoted by the notches of the boxplots) of the second and third trial. The change in average scores from trial 5 (individual) to trial 6 (collaborative) was the only change in average score that was statistically significant ($p = 0.05$) that supported our hypothesis. For the second design group, the average scores are closer to our hypothesized scores, especially after the second trial. Yet, none of the changes in average scores were statistically significant ($p < .05$).

The boxplots from Figure 6 show that the ninja model was the least popular of the three models, which can be especially appreciated in the first design group. In the first design group, the individually created robot model and the collaboratively created female model were considered the most creative. About 75% of the answers (as indicated by the top of the box) for these two models were concentrated below the median score of 4. In the second design group, the individually and collaboratively created female models were considered the most creative. Finally, the ninja model was considered the least creative in both the first and the second design group.

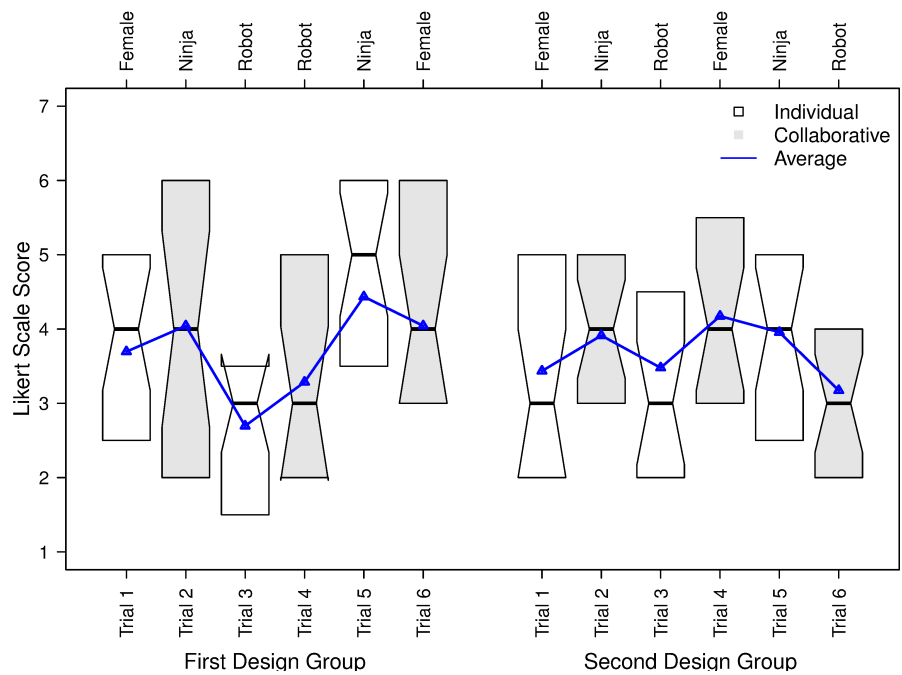

**Fig. 7** The transformation can be used in a video game (1 – very true, 7 – not true at all).

Figure 7 illustrates the scores for the statement of whether the transformation could be used in a video game. The white robot in both the individual and collaborative sessions of both design groups was found to be the most suitable to be used in a video game as shown by the lowest average scores. For trials 1-3 in the first and second design groups, the results are the opposite of our hypothesis, while for trials 4-6 the results are closer to our hypothesis. For the first design group, the change in average from trial 2 (collaborative) to trial 3 (individual) was statistically significant (p < .05), and the change in average from trial 4 to trial 5 was also statistically significant (p < .05), with the latter change supporting our hypothesis. None of the other changes in average scores were statistically significant in the first design group. In the second design group, none of the changes were statistically significant.

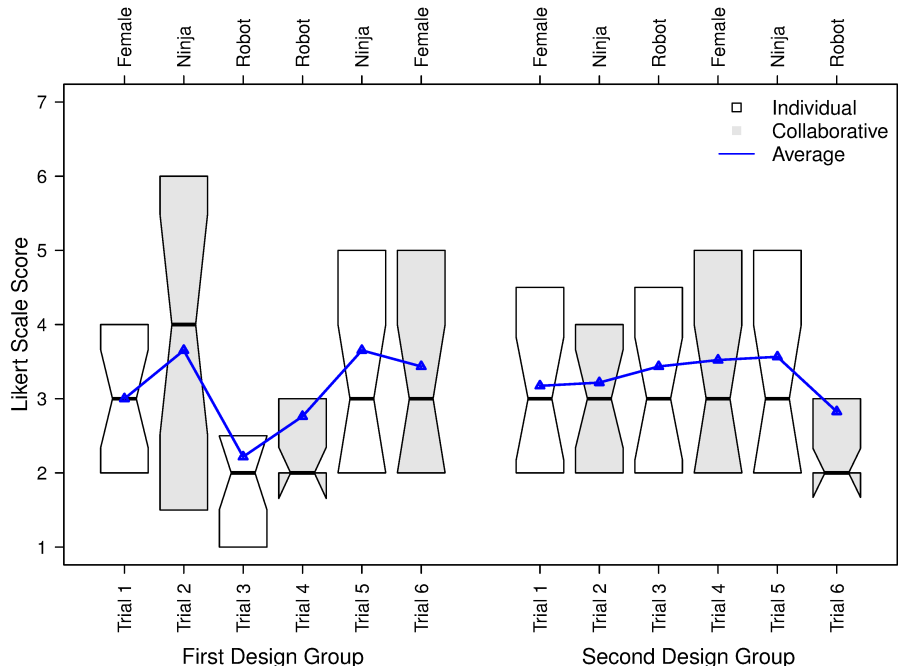

**Fig. 8** The transformation with minor tweaks can be used in a video game (1 – very true, 7 – not true at all).

Figure 8 shows the evaluation scores for the question asking whether the model with minor tweaks could be used in a video game. The model found to be most suitable after minor tweaks was the white robot, which is consistent with the results from Figure 7. In the first design group, the individually created robot received the best scores, whereas in the second design group the collaboratively created robot received the best scores. None of the changes in average scores were statistically significant (p < .05).

Figure 9 illustrates the evaluation scores for the novelty criterion. For the first and second design groups, the average scores and boxplots for trials 4-6 show the desired zig-zag pattern. However, for the first design group only the change in average score from trial 2 (collaborative) to trial 3 (individual) was statistically significant (p < .05). For the second design group, only the change from the trial 1 (individual) to trial 2 (collaborative) was statistically significant (p < .05). Both of these results are the opposite of our hypothesis.

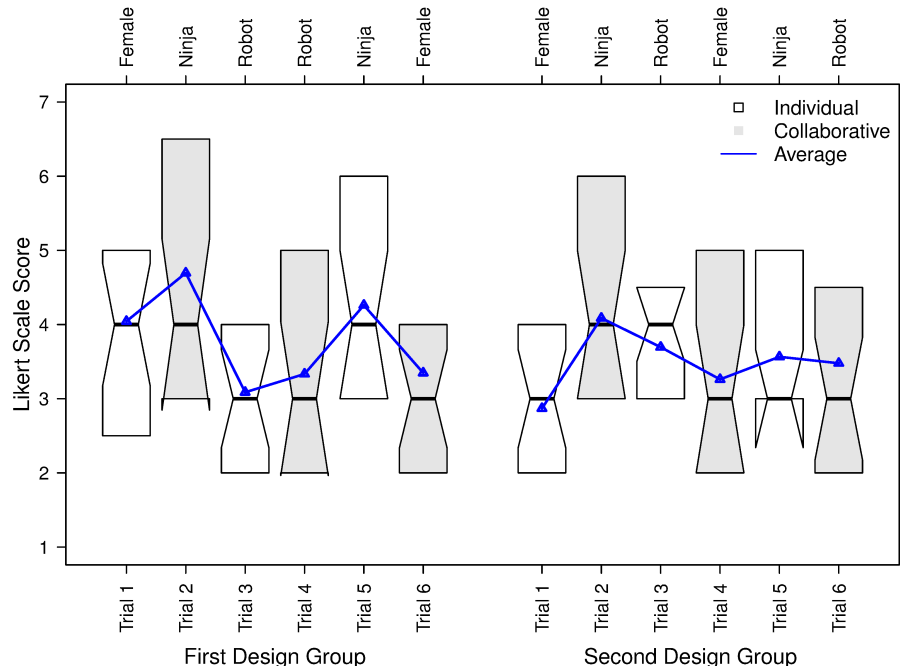

**Fig. 9** The transformation is novel (1 – very true, 7 – not true at all).

Figure 10 illustrates the scores for the surprising criterion. In the first design group, trials 4-6 reflect the desired score pattern. Furthermore, the collaboratively created blue female model received the best scores with at least 50% of scores being in the range 1-3. In the second design group, both the individually and collaboratively created solutions received similar scores. However, the collaboratively created blue female model received at least 25% of scores in the range 1-2.

For the second design group, none of the changes in average scores were statistically significant. For the first design group, the change from trial 2 to trial 3 was statistically significant (p < .05), which does not support our hypothesis. Yet, the change from trial 4 to trial 5, and from trial 5 to trial 6 were statistically significant (p < .05), while also supporting our hypothesis. This latter result is encouraging as it suggests that the 3D mod-els were considered more surprising when our collaborative model was used to generate model transformations.

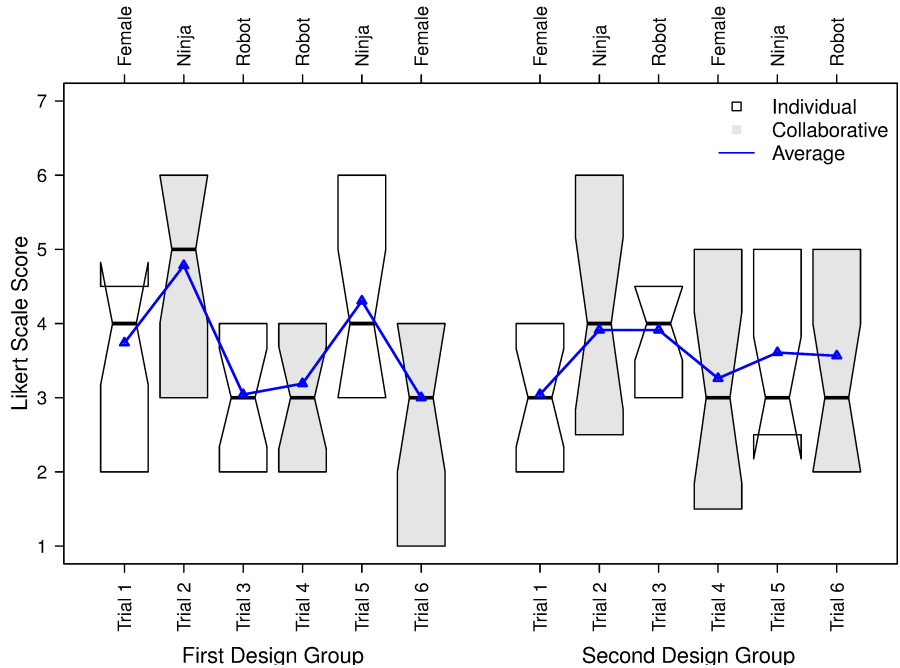

**Fig. 10** The transformation is surprising (1 – very true, 7 – not true at all).

## Online Evaluation

The first evaluation phase showed that models created individually received similar scores to models created collaboratively. We believed that it would be difficult to prove that every solution generated collaboratively would be more creative than a solution created individually. In view of the results from the first evaluation, we created an online evaluation in which users viewed a video and provided scores for the models in the video via an online survey.

The video first presented the three original models, without any transformations, so that participants could appreciate the differences that the transformations made. The video then showed a first row of individually created transformations, followed by a second row of collaboratively created transformations. The online survey used the same evaluation criteria from the first evaluation phase, with one exception. We removed the ques-tion asking whether the transformation with minor tweaks could be used in a video game to make the survey shorter. The survey first asked participants to evaluate the row of individually created models, followed by evaluation of the row of collaboratively created models. After both rows of models had been evaluated, participants were asked which of the two rows they liked the most and which of the two rows was the most creative.

We had a total of 16 completed online evaluations. We aggregated the scores for all of the individual models per criterion and similarly for the collaborative models. We compared the averages using the independent samples t-test and we found no statistically significant differences in the results. With regard to which row participants liked the most, eight participants picked the row of individually created models, seven participants picked the row of collaboratively created models, and one participant did not answer. With regard to which row was the most creative, three participants picked the row of individually created models and 13 participants picked the row of collaboratively created models.

**Discussion**

Figure 11 shows examples of two models generated by the design participants. All of the effects on the models involved animation. Thus, it was difficult to capture the resulting effects in images. Our observations were that while both individual and collaborative solutions were interesting, the collaborative solutions had more dramatic effects. In addition, the collaborative solutions tended to be more chaotic, and thus had a less polished look.

Figure 11(a) shows a blue female model created individually. The female model starts by standing straight, and the evolved vertex program made the body of the female model curve from left to right in the shape of an "S." Even though the model is curved, all parts of the original model can be identified, and overall the model has a smooth and aesthetically pleasing look. Figure 11(b) shows a blue female model created collaboratively. The evolved vertex program made the female model expand upwards, making the model look like the tail of a comet. The interesting part of the model is that in the midst of the chaotic animation, the face of the model remains visible, along with some of the extremities, such as both legs and part of her arms. Yet, this model, while exciting, does give the impression of being a work in progress.

From the results from the first evaluation phase we can deduce that models created individually were equally creative as models created col-laboratively. There are observations which we believe help explain some of the results obtained. First, we found that transformations on the blue female models resulted in the most interesting and aesthetically pleasing effects. The geometry and the skeleton, which dictates how a model moves when animated, of the female model resulted in transformations that looked polished and well-done compared to the transformations of the other two models. Transformations on the female model tended to have smooth transitions, and many times the original model was deformed in a curved fashion, resulting in soft edges. On the other hand, transformations on the robot and ninja models tended to result in sharp and jagged edges. In fact, we found that the same transformation applied to different models resulted in different effects due to the model geometry and the skeleton. Overall, the ninja models resulted in the less interesting effects. This was particularly identified while we were building the sets of best individual models and best collaborative models for the online evaluation.

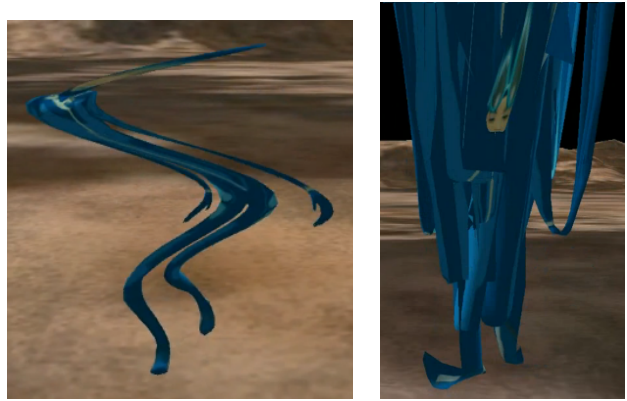

**Fig. 11** Individual (a) and collaboratively (b) generated transformations of the female model.

Regarding the online evaluation results, the fact that the majority of participants identified the row of collaborative models as the most creative, yet the average evaluation scores of collaborative and individual models were not statistically different, suggests a couple of points. First, the structure of the survey may have created a learning confound. In the online evaluation, the video was not included in the online survey web page. Participants had to follow a link to watch the video on an external web page. Therefore, participants would have had to switch back and forth between the survey web page and the video web page, instead of being able to score the models with the video always in view. A further nuance was that the video focused on each model for an average of 10 to 15 seconds. We had to impose this time limit in order to keep the video short. If participants wanted to watch one of the models in further detail, then they would have had to rewind the video to the right spot. Finally, we do not know whether participants watched the entire video once and then filled out the online survey, or whether they scored each model in tandem with the video. All of these nuances may have resulted in a learning confound that negatively biased the results.

**Conclusions**

In this paper we presented and evaluated the results of design evaluation on evolving a set of transformations on 3D models used in video games. The evolution is based on a collaborative model of creative design that uses interactive genetic algorithm and collaboration via solution injection among peers. The study is a major extension of our previous work in evolving 2D architectural floorplans. In the present study we look at issue of expanding the design space by introducing new variables, which has the potential to lead to uncovering solutions (creative or otherwise) that would not have been possible otherwise. While the evaluation scores did not fully support our hypothesis that our computational model of creative design increases creative content of solutions, the use of our collaborative model materially changed the resulting designs due to the expansion of the search space via collaboration. Finally, our work demonstrates that our collaborative IGA computational model matched with designer collaboration offers a valuable mixed-initiative approach to the use of evolutionary systems in design.

   As part of the experimental design, we limited how the search space could be expanded during the course of evolution. This is different than how creativity occurs in design practice, where a designer expands the search space as a result of a better understanding of the problem and solution, leading to a creative leap. Therefore, the model needs to be further validated by testing the use of the model on an actual design task. We foresee designers encoding different requirements to explore with evolution, letting designers explore solutions collaboratively with IGAs, and use this as a basis for discussion of potential design solutions. The work presented here can be thought of as the general framework of incorporating IGAs in design with peer-to-peer collaboration with the objective of evolving

"more creative" solutions than what is possible in a non-collaborative environment.